\documentclass[11pt]{article}

\usepackage[final]{acl}

\usepackage{times}
\usepackage{latexsym}

\usepackage[T1]{fontenc}
\usepackage[utf8]{inputenc}

\usepackage{microtype}

\usepackage{inconsolata}

\usepackage{graphicx}
\usepackage{enumitem}
\usepackage{algorithm}
\usepackage{algpseudocode}

\usepackage{algorithm}
\usepackage{algpseudocode}
\usepackage{amssymb}
\usepackage{amsmath}

\usepackage{microtype}           % nicer line breaking
\usepackage{xurl}                % lets \url break at more places
\usepackage{hyperref} % load near the end
\usepackage{booktabs} 
\usepackage{multirow} 

\DeclareMathOperator*{\argmax}{arg\,max}

\title{\includegraphics[width=0.9cm]{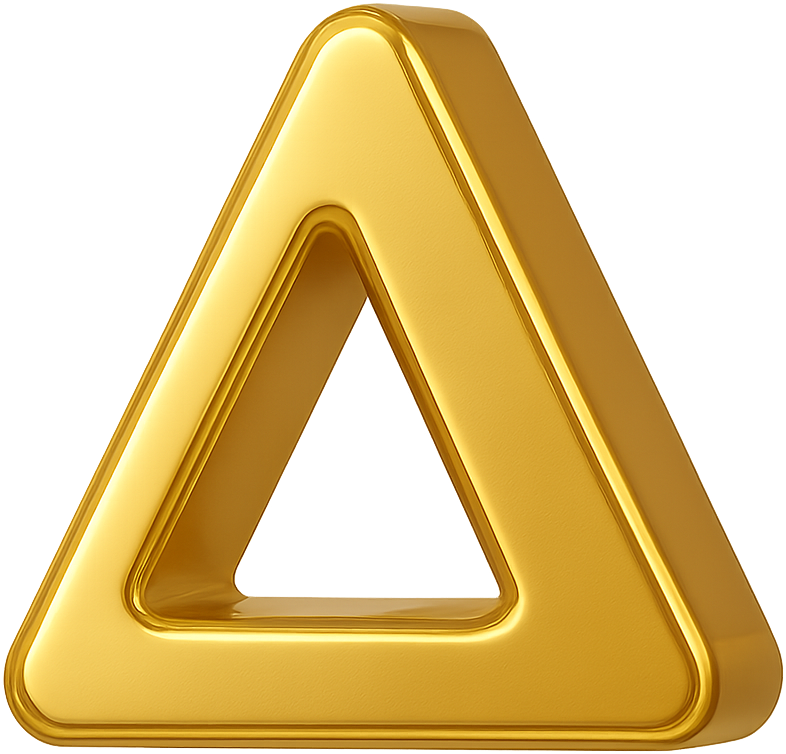} DELTA: Dynamic Layer-Aware Token Attention \\ for Efficient Long-Context Reasoning}

\author{%
  Hossein Entezari Zarch, Lei Gao, Chaoyi Jiang, Murali Annavaram\\
  University of Southern California\\
  \texttt{\{entezari, leig, chaoyij, annavara\}@usc.edu}\\
}

\begin{document}
\maketitle
\begin{abstract}
Large reasoning models (LRMs) achieve state-of-the-art performance on challenging benchmarks by generating long chains of intermediate steps, but their inference cost is dominated by decoding, where each new token must attend to the entire growing sequence. One approach to reduce this latency is to evict entries from the key-value (KV) cache, thereby reducing the active context used in attention computation. However, such sparse attention methods suffer from severe accuracy degradation on reasoning tasks due to cumulative selection errors and the evolving importance of tokens over long derivations. We present \textbf{DELTA}, a training-free sparse attention mechanism that improves computational efficiency without sacrificing model accuracy. DELTA partitions transformer layers into three groups: initial layers that use full attention, a small set of \emph{$\Delta$-layers} that identify salient tokens via aggregated head-level attention scores, and subsequent sparse-attention layers that attend only to the selected subset. This design preserves the full KV cache in GPU memory for accuracy, while avoiding expensive full-attention computation over many layers. On reasoning benchmarks such as AIME and GPQA-Diamond, DELTA matches or surpasses full attention in accuracy, while reducing the number of attended tokens by up to $4.25\times$ and delivering $1.54\times$ end-to-end speedup. Our results show that selective reuse of intermediate attention maps offers a robust path toward efficient long-context reasoning. The code is available at \url{https://github.com/hoenza/DELTA}.
\end{abstract}

\section{Introduction}
Recent progress in large language models (LLMs) has led to systems with impressive capabilities in reasoning and self-reflection. Large reasoning models (LRMs) such as DeepSeek-R1 \citep{guo2025deepseek}, Gemini-2.5-pro \citep{google2025gemini25}, OpenAI-o3 \citep{openai2025o3o4mini}, Qwen3 \citep{yang2025qwen3}, and GPT-OSS \citep{openai2025gptoss} leverage test-time scaling by generating long chains of intermediate reasoning steps, significantly improving accuracy on challenging benchmarks \citep{codeforcesamerican, rein2024gpqa, hendrycks2021measuring, wei2022chain}. However, serving such models efficiently remains difficult due to severe memory and compute bottlenecks in attention operation \citep{vaswani2017attention}, especially under long-context generation settings \cite{dao2023flashattention, ye2025flashinfer}.

LLM inference consists of two stages: \emph{prefilling} and \emph{decoding}. In the prefilling stage, the model processes the prompt, computes hidden representations, and materializes all key–value (KV) vectors as a KV cache in GPU high-bandwidth memory (HBM). During decoding, tokens are generated autoregressively: for each new token, the model computes its KV vectors, appends them to the cache in HBM, and attends over the entire history to produce the next output. Because the KV cache grows linearly with sequence length and batch size \citep{kwon2023efficient}, the amount of data that must be read from HBM increases rapidly. For instance, with a 32K-token context and a batch size of 128, the KV cache of Llama-3-8B in float16 already exceeds 500 GB.\footnote{Computed as Layers × Sequence Length × Batch Size × KV Heads × Head Dim × 2 (for K\&V) × 2 bytes = $32 \times 32\text{K} \times 128 \times 8 \times 128 \times 2 \approx 512$ GB.} Unlike the prefilling stage, which writes the KV cache once, the decoding stage must repeatedly stream all previously stored KV entries from HBM for every new token. This makes decoding inherently \emph{memory-bandwidth bound}: throughput is limited by the cost of moving hundreds of gigabytes of KV data per step. As context length or batch size grows, this bandwidth pressure scales linearly, quickly overwhelming GPU memory systems and severely constraining long-context inference.

These bandwidth limitations are particularly acute for reasoning workloads. Unlike typical NLP tasks that involve long inputs but short outputs, reasoning problems often begin with concise prompts yet require lengthy derivations spanning tens of thousands of tokens. This decode-heavy profile magnifies the bandwidth bottleneck, as each step involves scanning ever-larger KV caches. As a result, the decoding stage dominates both latency and resource usage: for example, using full attention in HuggingFace, DeepSeek-R1-Distill-Llama-8B requires more than 15 minutes on a single NVIDIA A100 GPU to generate 32K tokens for one AIME problem \citep{yue2025don}. Optimizing the decoding stage is therefore essential for efficient LLM serving in reasoning applications.

Meanwhile, the unique structure of reasoning workloads opens new opportunities for efficiency. While prefilling benefits from full attention to capture global context, the much longer decoding phase is well-suited to sparsity. Sparse attention reduces computation and bandwidth requirements by restricting reads to a subset of salient tokens rather than scanning the entire KV cache. Prior work has explored two complementary directions: selection-based methods \citep{tang2024quest, hao2025omnikv, liu2024retrievalattention, gao2025seerattention, yuan2025native, yang2024tidaldecode}, which preserve the full KV cache but attend only to chosen tokens, and eviction-based methods \citep{hu2025raas, li2024snapkv, xiao2023efficient, zhang2023h2o, adnan2024keyformer, cai2025r}, which permanently discard unselected tokens to reduce storage cost of KV cache. Both rely on identifying important tokens using predefined criteria, and together they demonstrate the potential of sparse attention as a foundation for efficient long-decode inference.

However, applying sparse attention to long reasoning generations remains challenging. Unlike standard generation tasks, where some information loss can be tolerated, step-by-step reasoning demands that critical context be preserved throughout the entire derivation to maintain logical consistency \citep{hu2025raas}. In practice, accuracy drops sharply when token selection errors accumulate over long sequences \citep{gao2025seerattention}. Eviction-based methods such as RaaS \citep{hu2025raas} illustrate this issue: by permanently removing tokens judged less important, they risk discarding tokens that later become essential once the generation length grows beyond the KV cache capacity. The core difficulty is twofold: (1) attention patterns evolve over time, and (2) a token’s importance can change, so tokens that seem irrelevant early may become highly influential later in the reasoning process.

At the same time, we make two key observations. First, attention maps across consecutive layers exhibit strong correlation: within a local block of layers, the first layer often predicts the important tokens for subsequent layers with high reliability. Second, attention distributions change gradually during decoding, which suggests that token importance can be predicted using intermediate layers without computing full attention everywhere. These insights highlight both the risk of aggressive eviction and the opportunity for accurate, low-cost selection.

To address the accuracy–efficiency tradeoff, we introduce \textbf{DELTA}, a training-free, selection-based sparse attention mechanism. DELTA preserves the full KV cache but restricts computation to a carefully chosen subset of tokens at each decoding step. It operates as a plug-and-play module that leverages the full attention maps of a small set of intermediate layers to predict the salient tokens for the upcoming layers. By selecting tokens using head-wise attention signals together with a stable recency window, DELTA significantly reduces the runtime cost of attention without incurring noticeable accuracy degradation.
In summary, our contributions are:
\begin{itemize}[leftmargin=1em]
    \item We provide a detailed token-level analysis of attention distributions in large reasoning models, revealing two properties: (1) strong correlation of attention patterns across consecutive layers, and (2) gradual but ongoing shifts in token importance during long generations.
    \item We propose \textbf{DELTA}, a training-free sparse attention mechanism that combines a head-aware token scoring rule with a stable recency window to retain the recent context most critical for reasoning.
    \item We demonstrate that DELTA achieves accuracy on par with, or better than, full attention on challenging reasoning benchmarks, while delivering up to $1.54\times$ end-to-end speedups. Compared with state-of-the-art sparse attention methods, DELTA reduces the number of attended tokens by up to $4.25\times$, all without sacrificing accuracy.
\end{itemize}

\section{Background}
\textbf{LLM inference process.}
Decoder-only LLMs generate tokens auto-regressively in two stages: the prefilling stage and the decoding stage. 

\paragraph{Prefilling stage.}  
At layer $i$, the input hidden states are $X^i \in \mathbb{R}^{b \times s \times h}$, where $b$ is the batch size, $s$ is the prompt length, and $h$ is the embedding dimension. Queries, keys, and values are projected as
\begin{equation}
Q^i = X^i W^i_Q; K^i = X^i W^i_K; V^i = X^i W^i_V
\label{eq:gqa_projection}
\end{equation}
where
\[
W^i_Q \in \mathbb{R}^{h \times h}, \quad 
W^i_K, W^i_V \in \mathbb{R}^{h \times (g \cdot d_{\text{head}})}.
\]
Here $m$ denotes the number of query heads, $g \leq m$ the number of KV groups, and $d_{\text{head}} = h/m$ the per-head dimension.

In grouped-query attention (GQA), the queries are divided into $m$ query heads,
\begin{equation}
Q^i = [Q^i_1, \ldots, Q^i_m], \quad Q^i_j \in \mathbb{R}^{b \times s \times d_{\text{head}}},
\end{equation}
while the keys and values are divided into only $g$ groups,
\begin{equation}
\begin{aligned}
K^i&=[K^i_1,\ldots,K^i_g],\quad V^i=[V^i_1,\ldots,V^i_g],\\
&K^i_\ell,V^i_\ell\in\mathbb{R}^{b\times s\times d_{\mathrm{head}}}.
\end{aligned}
\end{equation}
Each query head $j$ is assigned to one KV group $\phi(j) \in \{1,\ldots,g\}$.
GQA generalizes standard attention mechanisms: when $g = m$, it reduces to multi-head attention (MHA), and when $g = 1$, it reduces to multi-query attention (MQA).

The scaled dot-product attention for head $j$ is
\begin{equation}
A^i_j = \frac{Q^i_j (K^i_{\phi(j)})^\top}{\sqrt{d_{\text{head}}}}, 
\quad O^i_j = \mathrm{softmax}(A^i_j) V^i_{\phi(j)},
\label{eq:gqa_attention}
\end{equation}
where $A^i_j$ are the attention scores and $O^i_j \in \mathbb{R}^{b \times s \times d_{\text{head}}}$ is the head output.

The outputs of all query heads are concatenated and linearly projected:
\begin{equation}
O^i = [O^i_1, \ldots, O^i_m] W^i_O, \quad W^i_O \in \mathbb{R}^{h \times h}.
\label{eq:gqa_output}
\end{equation}

A feed-forward network (FFN) follows the GQA block:
\begin{equation}
X^{i+1} = \sigma(O^i W^i_1) W^i_2,
\label{eq:ffn}
\end{equation}
where $W^i_1 \in \mathbb{R}^{h \times d_{\text{FFN}}}$, $W^i_2 \in \mathbb{R}^{d_{\text{FFN}} \times h}$, and $\sigma(\cdot)$ is a non-linear activation.

\paragraph{Decoding stage.}  
At step $t$, each layer receives a single token embedding $x^i \in \mathbb{R}^{b \times 1 \times h}$. The new key and value are concatenated to the cached ones:
\begin{equation}
K^i \leftarrow [\,K^i;\; x^i W_K^i\,], V^i \leftarrow [\,V^i;\; x^i W_V^i\,].
\label{eq:kv_update}
\end{equation}
The subsequent GQA and FFN computations mirror the prefilling stage.

\paragraph{Decode cost and memory I/O.}\;
While prefilling writes the KV cache once, decoding must repeatedly read all past K/V entries for each new token, making long-context inference inherently memory-bandwidth bound. Prior work reports that decoding dominates end-to-end latency under long contexts and that KV memory movement constitutes a major fraction of decode time, underscoring the need to reduce KV reads without sacrificing accuracy \citep{kwon2023efficient,dao2023flashattention}.

\begin{figure*}[t!]
    \centering
    \includegraphics[width=1\linewidth]{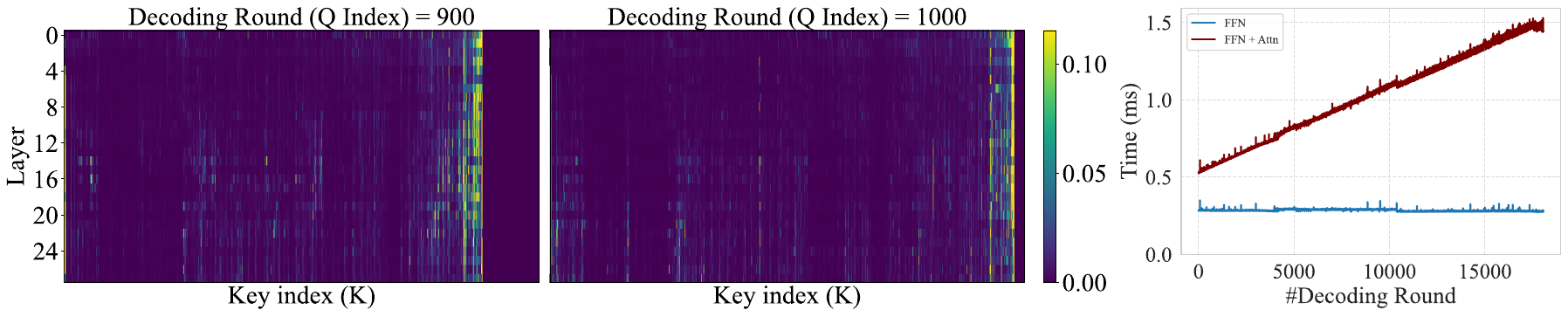}
    \caption{
    (Left) Attention maps from Qwen-7B at decoding steps 900 and 1000, where each row corresponds to a Transformer layer.
    (Right) Decoding runtime of FFN and attention modules across generation, showing attention’s linear growth with context length.}
    \label{fig:observation1}
\end{figure*}
\paragraph{Sparse attention.}
Full attention requires each query to attend to all past tokens, which scales linearly with sequence length. For clarity, the sparse-selection formulation below is written for a single decoding query, and we therefore omit the batch dimension. Sparse attention reduces this cost by restricting computation to a subset of $k$ tokens. For head $j$, let the exact attention weights be
\[
\alpha^i_j = \mathrm{softmax}(A^i_j) \in \mathbb{R}^s.
\]
Instead of attending to all $s$ tokens, we select an index set $\rho \subseteq \{1,\ldots,s\}$ with $|\rho|=k$. Since computing $\rho$ from $\alpha^i_j$ directly is expensive, practical methods rely on an approximation function $f$ that predicts which tokens are likely to have high attention:
\begin{equation}
\rho = \argmax_{\rho' : |\rho'| = k} f(Q^i_j, K^i_{\phi(j)}, V^i_{\phi(j)}, \rho').
\label{eq:subset_selection}
\end{equation}

The quality of the selection is measured by the \emph{attention recall}, defined as the fraction of the ground-truth attention mass preserved in the selected subset:
\begin{equation}
R^i_j = \frac{\sum_{u \in \rho} \alpha^i_j(u)}{\sum_{u=1}^s \alpha^i_j(u)}.
\label{eq:recall}
\end{equation}
Maximizing $R^i_j$ under the budget constraint $k$ is the central objective of sparse attention methods, ensuring efficiency while maintaining accuracy.

\paragraph{Sparsity and query dependence.}\
Self-attention exhibits substantial sparsity beyond the earliest layers: a small subset of critical tokens typically accumulates most attention mass, enabling accurate computation on a reduced context. However, criticality is strongly \emph{query dependent}: the tokens that matter vary with the current query vector $Q$, and may change rapidly across consecutive decode steps. Heuristics based only on past usage (eviction) risk losing later-salient tokens, whereas query-aware selection retains high recall under long reasoning traces \citep{tang2024quest,zhang2023h2o,ge2023model}.

\section{Motivating Observation}
\label{sec:motivate}
\paragraph{Depth-wise context sharpening.}
Figure~\ref{fig:observation1} (left) illustrates how attention patterns evolve with depth. In early layers, the model primarily attends to nearby tokens and exhibits diffuse, low-mass attention over the broader context, showing little focus on distant information. As depth increases, attention becomes progressively sharper and more selective, concentrating on a small set of far-away tokens that carry high relevance.

\paragraph{Layer-wise correlation.}
Empirical profiling of large reasoning models such as Qwen-7B reveals that consecutive layers exhibit highly correlated attention patterns.
Tokens that receive high attention in one layer tend to remain salient in the next layers, as illustrated in Figure~\ref{fig:observation1} (left), which visualizes layer-wise attention maps at decoding steps 900 and 1000 of a reasoning sequence.
Each row corresponds to a Transformer layer, showing that deeper layers largely preserve the spatial configuration of attention established in earlier layers.
This structural continuity suggests that adjacent layers refine rather than reconstruct attention, enabling later layers to reuse the relational patterns captured by their predecessors.
As a result, computing full attention in every layer becomes redundant: once salient tokens are identified, subsequent layers can effectively operate on a reduced, high-recall subset of the context.

\paragraph{Sequential drift.}
While the overall structure of attention is stable across depth, it evolves gradually along the decoding trajectory.
Between decoding steps 900 and 1000 in Figure~\ref{fig:observation1}, the regions of strongest attention shift across key positions, revealing how the model dynamically repositions its focus as new tokens are generated.
We refer to this phenomenon as \emph{sequential drift}.

This progressive movement reflects an adaptive retrieval process, where the model continuously updates which parts of the context are relevant to the current query embedding $Q_t$.
Such behavior highlights the need for a \emph{query-adaptive} sparse attention mechanism that dynamically adjusts token selection at each decoding step rather than relying on fixed or history-based heuristics.

\paragraph{Runtime dominated by attention.}
Figure~\ref{fig:observation1} (right) presents the measured decoding runtimes of FFN and attention modules.
While the FFN cost remains nearly constant, attention latency increases almost linearly with context length.
Beyond 8k tokens, attention dominates total inference time, driven primarily by repeated KV-cache memory access rather than compute operations.
These trends confirm that long-context decoding is bottlenecked by attention and motivate our approach: performing full attention only in a few strategically chosen layers to identify salient tokens, while letting the remaining layers operate on a compact, high-recall context subset.

\section{DELTA: Dynamic Layer-Aware Token Attention}
\begin{figure*}[t!]
    \centering
    \includegraphics[width=1\linewidth]{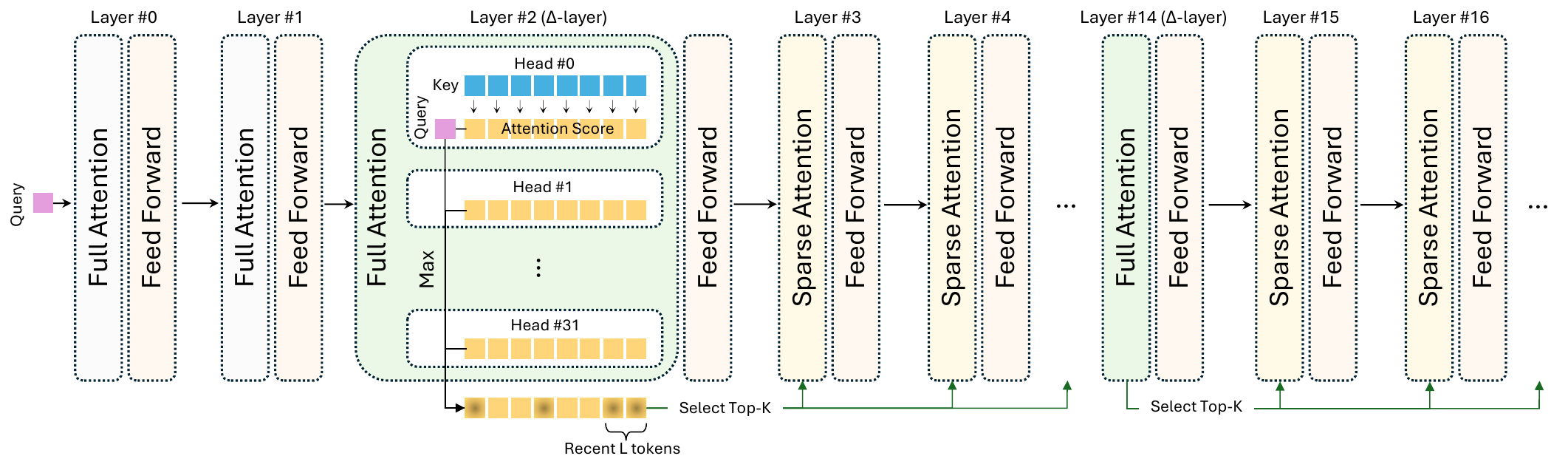}
    \caption{Overview of the DELTA decoding process. The first two layers perform full attention for initialization, $\Delta$-layers (e.g., Layers 2 and 14) run full attention to select salient tokens, and subsequent sparse attention layers attend only to those selected tokens, as indicated by green arrows.}
    \label{fig:delta}
\end{figure*}
The empirical patterns described in Section~\ref{sec:motivate} motivate a layer-aware sparse attention design that minimizes redundant computation while preserving reasoning accuracy. In early layers, attention maps are diffuse and unstable, requiring full-sequence attention to build reliable representations. In contrast, deeper layers exhibit high inter-layer correlation: once a small set of context tokens becomes salient, subsequent layers largely reuse them. Finally, as decoding proceeds, the regions of strong attention shift gradually along the sequence, a phenomenon we term \emph{sequential drift}. Together, these observations suggest that only a few layers need to compute full attention to refresh salient tokens, while the rest can reuse them at low cost.

\paragraph{Core idea.}
DELTA operationalizes this principle through a structured three-tier layer design. The first few layers perform full attention to stabilize representations, as early layers show no consistent sparsity structure. A small number of intermediate $\Delta$-layers act as \emph{selection layers} that re-run full attention to identify a compact set of salient tokens carrying most of the attention mass. The remaining layers perform sparse attention restricted to those selected tokens, reusing them until the next $\Delta$-layer updates the selection. This design removes redundant full-sequence computation across correlated layers while maintaining a high-recall context for reasoning.

\paragraph{Query-adaptive refresh.}
Because attention focus evolves with each new query embedding $Q_t$, the salient set must be updated at every decoding step. Skipping or caching old selections would cause stale focus and recall loss. Therefore, each $\Delta$-layer recomputes full attention per generated token, ensuring that the reduced context remains aligned with the evolving query. However, this update occurs only at the sparse set of $\Delta$-layers, keeping total computation cost low. Notably, DELTA never discards tokens from the KV cache: older tokens may regain relevance due to long-range reasoning dependencies. Instead, it restricts computation, not memory, by letting sparse attention layers attend only to the currently selected subset while retaining the full cache for later refreshes.

\paragraph{Token-level formulation.}
We first describe DELTA as selecting salient tokens under a fixed token budget. This formulation clarifies the selection objective. We then describe the page-based implementation used in practice for efficient KV-cache access on GPUs.

\paragraph{Selection mechanism.}
At each $\Delta$-layer $i$, we form a reduced token set $\rho$ by preserving a recency window of $\ell$ tokens and selecting the remaining $k-\ell$ tokens by importance. Let $\alpha^i_j=\mathrm{softmax}(A^i_j)\in\mathbb{R}^{s}$ denote the attention weights of head $j$ (Eq.~\ref{eq:gqa_attention}). We define the score of token $t$ as its maximum attention weight across heads:
\[
s_t = \max_{j=1,\dots,m}\alpha^i_j(t).
\]
We then select the top-$(k-\ell)$ tokens by $s_t$ among $\{1,\dots,s-\ell\}$ and merge them with the recency window:
\[
\rho=\operatorname{Topk}\big(\{s_t:t\le s-\ell\},\,k-\ell\big)\cup\{s-\ell+1,\dots,s\}.
\]

\paragraph{Page-based DELTA.}
Token-level KV management fragments memory and hinders efficient GPU access. Following common practice, we store the KV cache in fixed-size pages of $P$ tokens \citep{kwon2023efficient,dao2023flashattention}. For the page-based implementation, let $K$ denote the total page budget and $L$ the recency-window budget in pages; the corresponding token budgets are $k=KP$ and $\ell=LP$. Let $\mathcal{P}=\{1,\dots,\lceil s/P\rceil\}$ be the page set and let $p(t)\in\mathcal{P}$ map token $t$ to its page. We score each page by summing the token scores it contains:
\[
S_u = \sum_{t:\,p(t)=u} s_t.
\]
We then keep the last $L$ pages and select the top-$(K-L)$ remaining pages by $S_u$. The reduced context is the union of tokens in these pages, enabling coalesced memory access and lower gather/scatter overhead during decoding. All experiments in this work use the page-based implementation.

\section{Experiments}
\paragraph{Experimental setup.}
We evaluate DELTA on four distilled DeepSeek-R1 variants \citep{guo2025deepseek}: DeepSeek-R1-Distill-Qwen-1.5B, 7B, and 14B, and DeepSeek-R1-Distill-Llama-8B, abbreviated as DS-Qwen-1.5B, DS-Qwen-7B, DS-Qwen-14B, and DS-Llama-8B, respectively. We assess reasoning performance on four open-source benchmarks: AIME-2024, AIME-2025 \citep{codeforcesamerican}, GPQA-Diamond \citep{rein2024gpqa}, and MATH500 \citep{hendrycks2021measuring}. The AIME datasets contain advanced high-school mathematics problems spanning algebra, geometry, number theory, and combinatorics. MATH500 is drawn from high-school competitions and covers five difficulty levels in the Art of Problem Solving framework, while GPQA-Diamond evaluates graduate-level scientific reasoning across biology, chemistry, and physics. We evaluate all 30 problems from AIME-2024 and AIME-2025, and the first 100 problems from GPQA-Diamond and MATH500.

\paragraph{Implementation details.}
We implement DELTA with the FlashInfer Just-In-Time (JIT) module \citep{ye2025flashinfer}, which extracts attention logits directly from the decoding kernel, and use the native PyTorch \texttt{topk} operator for page selection. This design keeps DELTA lightweight and easy to integrate across models. Unless otherwise specified, all experiments use the page-based implementation of DELTA with page size $P$$=$$16$, following common practice \citep{kwon2023efficient,dao2023flashattention}. For readability, we sometimes report budgets in token-equivalent units; for example, a page budget of $K$$=$$64$ corresponds to 1k tokens. We reproduce the baselines, RaaS and Quest, in Hugging Face Transformers \citep{huggingface2025} to ensure a consistent comparison. All experiments are conducted on a single node with eight NVIDIA A100 (SXM4, 40GB) GPUs.

\begin{figure*}[t!]
    \centering
    \includegraphics[width=1\linewidth]{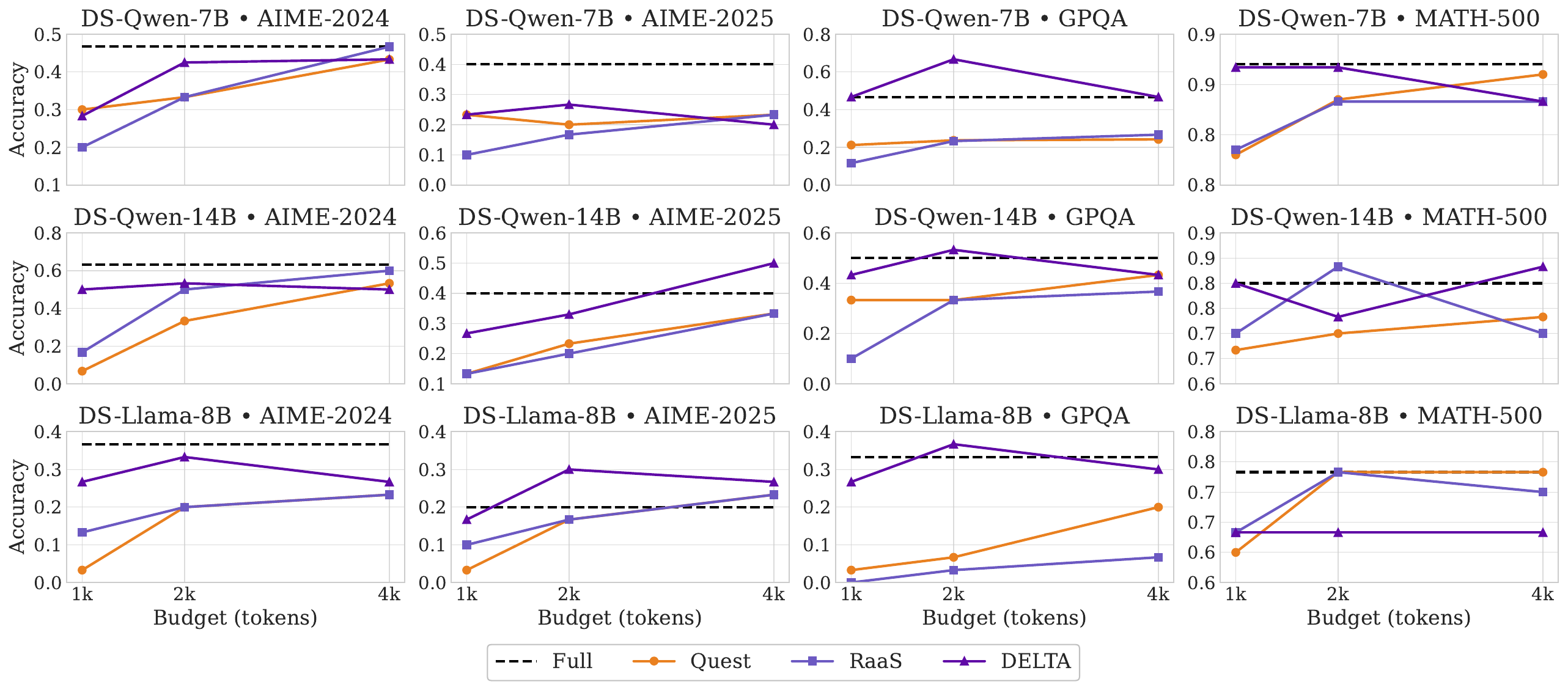}
    \caption{Accuracy of sparse attention methods on reasoning benchmarks. DELTA consistently matches or exceeds the accuracy of full attention under limited context budgets and remains robust across datasets.}
    \label{fig:result1}
\end{figure*}

\paragraph{Baselines.}  
We compare DELTA against three approaches. \textbf{Full} denotes standard decoding where all layers attend to the entire KV cache. \textbf{Quest} \citep{tang2024quest} is a selection-based method that compresses each KV page into two representative vectors (element-wise min/max of keys), scores pages against the current query, and retrieves the top-$k$ for attention; it preserves the full cache in HBM but incurs overhead from storing representatives (two key vectors per page, i.e., $1/8$ of KV memory for a page size of 16). \textbf{RaaS} \citep{hu2025raas} is an eviction-based method that removes pages with consistently low attention scores, lowering memory usage but risking the loss of tokens that may later become important.

\paragraph{Metrics.}
\emph{Accuracy} measures whether the model’s final answer is mathematically equivalent to the ground-truth answer, and is reported as the fraction of correctly solved problems in the evaluation set. \emph{Decoding length} denotes the number of tokens generated before either the end-of-sequence token or the maximum generation limit is reached, capturing the length of the reasoning trajectory. \emph{Throughput} is defined as the total number of generated tokens divided by the total decoding time. \emph{Forward time} is defined as the time required for a single model forward pass.

\subsection{$\Delta$-layer configuration}
We use a fixed $\Delta$-layer configuration for each model across all experiments, unless otherwise specified. For all models, we use full attention in layers $[0,1]$ during initialization, since the earliest layers exhibit diffuse attention and do not yet form stable sparsity patterns. Layer $[2]$ is always selected as the first $\Delta$-layer to perform the initial salient-page selection. Two additional $\Delta$-layers are placed later in the network to refresh the selected pages as attention evolves with depth.

\paragraph{$\Delta$-layer calibration.}
To choose these later $\Delta$-layers, we use a lightweight calibration procedure on a small calibration set. Specifically, we run full-attention decoding, record the attention maps of all layers, and compute the average inter-layer shift between consecutive layers using $1-\mathrm{cosine\ similarity}$, averaged over decoding steps and samples. We then select layers with large shifts while ensuring that they are well distributed across the network depth. The intuition is that large inter-layer shifts indicate transition points where the previous layer becomes less reliable for predicting salient pages for later layers, making these layers effective refresh points.

\paragraph{Model-specific $\Delta$-layers.}
Following this procedure, DS-Qwen-1.5B uses layers $[2,14,23]$ out of $[0\text{-}27]$, and DS-Qwen-7B uses layers $[2,14,22]$ out of $[0\text{-}27]$. DS-Qwen-14B uses layers $[2,6,42]$ out of $[0\text{-}47]$, and DS-Llama-8B uses layers $[2,8,31]$ out of $[0\text{-}31]$ as $\Delta$-layers. We use the same calibrated layer configuration for all datasets evaluated for a given model.

\subsection{Accuracy Results}
For all accuracy experiments, we use a page budget of $K$$=$$64$ with page size $P$$=$$16$, corresponding to a total token budget of 1k, and reserve $L$$=$$8$ pages (128 tokens) for the recency window. Figure~\ref{fig:result1} compares the accuracy of Quest, RaaS, and DELTA, across four reasoning datasets and three models. Three consistent patterns emerge. First, with a 1k-token budget, DELTA consistently outperforms existing sparse methods and often matches or even surpasses the full attention baseline. For instance, on \textsc{AIME-2024} with DS-Qwen-14B, Quest and RaaS achieve below 20\% accuracy, whereas DELTA attains nearly 50\%, approaching the 60\% accuracy of full attention. Second, increasing the token budget from 1k to 2k often improves performance, reflecting the diminishing effect of sparsity-induced selection errors. In several cases, DELTA with a 2k-token budget even surpasses the full attention baseline; for example, on \textsc{GPQA} with DS-Qwen-7B, DELTA outperforms full attention by roughly 30\%. Finally, expanding the budget further to 4k yields marginal or no improvement and occasionally a slight decline in accuracy. This plateau suggests that DELTA captures most salient context within small budgets, beyond which additional tokens primarily introduce redundancy rather than useful information.

\subsection{Speedup Results}
\begin{figure*}[t!]
    \centering
    \includegraphics[width=1\linewidth]{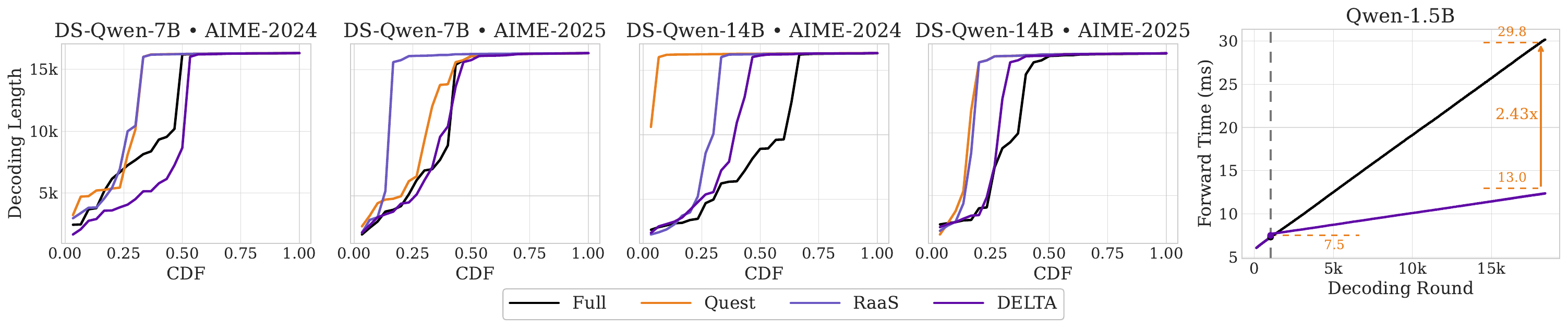}    
    \caption{(Left) CDF of decoding lengths across evaluated samples, where DELTA achieves shorter or comparable decoding lengths than full attention and other sparse-KV baselines. (Right) Once token selection is activated (gray line), DELTA slows the growth of latency relative to full attention, increasing throughput.}
    \label{fig:result2}
\end{figure*}

Figure~\ref{fig:result2} (left) shows the cumulative distribution function (CDF) of decoding lengths across evaluated samples, with the CDF on the x-axis and decoding length on the y-axis. Better methods achieve the same cumulative fraction at smaller decoding lengths, corresponding to curves that are shifted to the right. DELTA consistently matches or improves over full attention and outperforms other sparse-KV baselines, showing that its sparsity does not lengthen reasoning trajectories.

To measure runtime and throughput, we use synthetic decoding traces that increase the generated length from 1 token up to the target maximum length for each model. Figure~\ref{fig:result2} (right) shows the per-round decoding latency of full attention and DELTA with a 1k-token budget. Experiments are conducted on DS-Qwen-1.5B with batch size 64 and a maximum decoding length of 18k tokens. The gray vertical line marks the point where DELTA begins page selection. Beyond this point, latency grows much more slowly under DELTA: full attention increases from 7.5\,ms to about 30\,ms, whereas DELTA rises to only 13\,ms, corresponding to about 4$\times$ smaller growth. Overall decoding time decreases from 403 to 261 seconds, while throughput increases from 2,921 to 4,517 tokens/s, a 55\% improvement. These results show that DELTA improves end-to-end decoding efficiency without increasing output length.

\section{In-depth Analysis}
Unless otherwise specified, all experiments in this section use a mixed evaluation set of 120 samples, constructed from 30 samples from each benchmark: AIME-2024, AIME-2025, GPQA, and MATH500. We refer to this set as \textbf{Mixed120}.

\subsection{Effect of $\Delta$-layers Configuration}
\begin{figure}[htbp]
    \centering
    \includegraphics[width=\columnwidth]{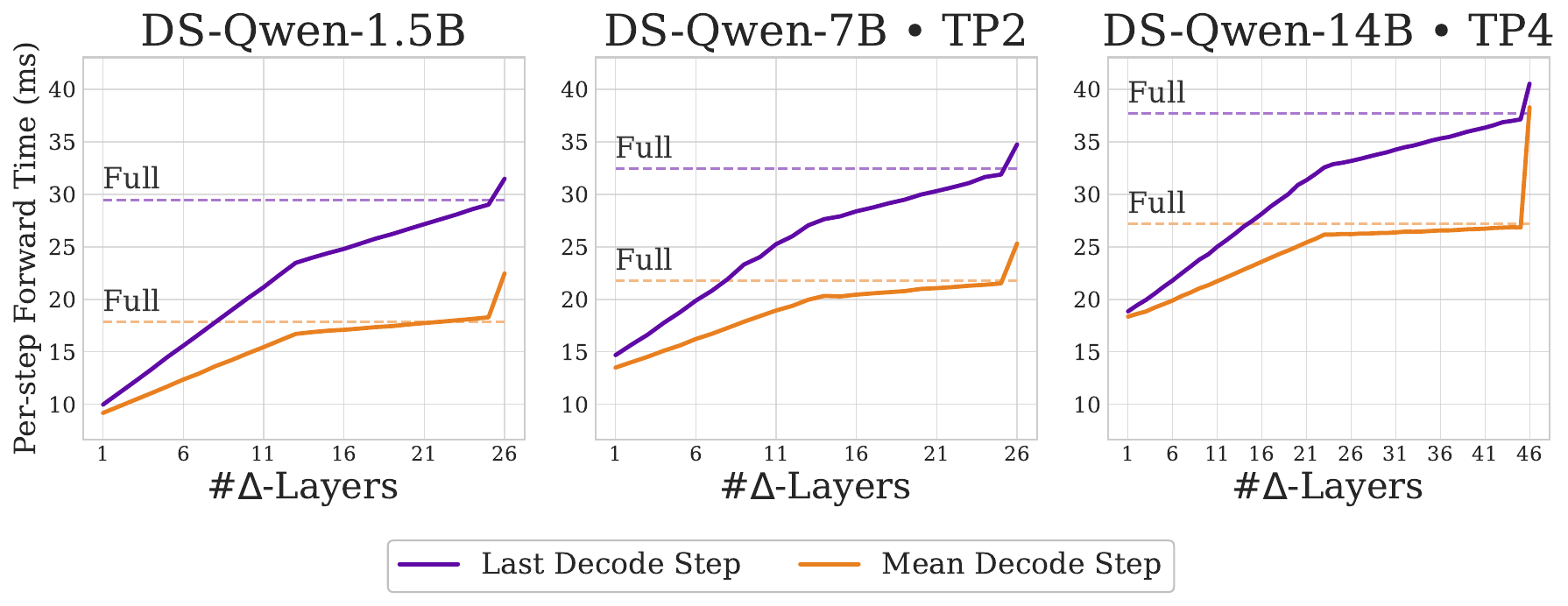}
    \caption{Adding more $\Delta$-layers consistently increases runtime and approaches the full attention baseline.}
    \label{fig:ablation1}
\end{figure}

We first study how the number of $\Delta$-layers affects runtime. We vary the number of $\Delta$-layers from 1 up to the maximum possible value. Figure~\ref{fig:ablation1} shows the results for DS-Qwen-1.5B, DS-Qwen-7B, and DS-Qwen-14B. DS-Qwen-1.5B is evaluated on a single GPU with batch size 64 and generation up to 18k tokens, DS-Qwen-7B on 2 GPUs with tensor parallelism and batch size 64 up to 16k tokens, and DS-Qwen-14B on 4 GPUs with tensor parallelism and batch size 32 up to 19k tokens. We report both the forward time of the last decoding step, corresponding to the maximum context length, and the average forward time across all decoding steps. The dashed lines indicate the corresponding full attention latencies. Across all models, increasing the number of $\Delta$-layers consistently increases runtime. This is because each additional $\Delta$-layer performs full attention to refresh the selected pages. As the number of $\Delta$-layers approaches the total number of layers, DELTA gradually reduces to full attention, and its runtime correspondingly approaches the full attention baseline.

\begin{figure}[htbp]
    \centering
    \includegraphics[width=\columnwidth]{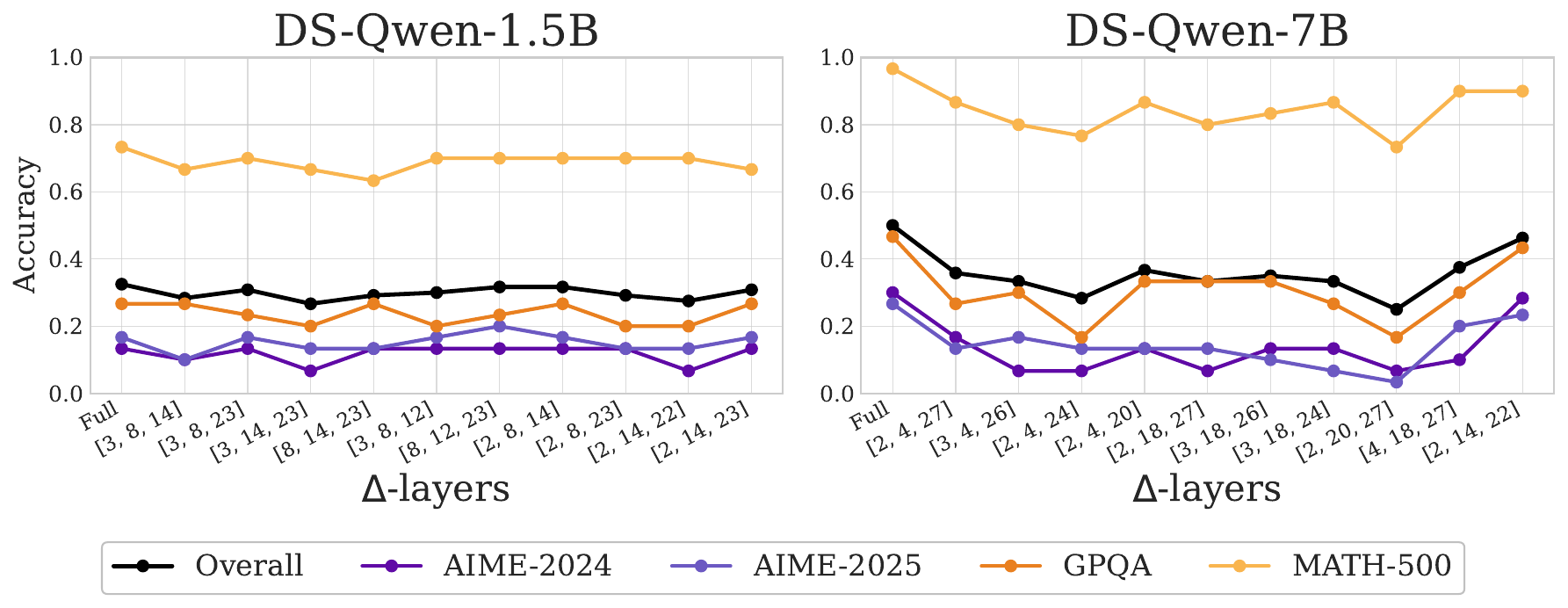}
    \caption{While overall accuracy varies across configurations, the trends across datasets remain largely consistent within each model.}
    \label{fig:ablation2}
\end{figure}

\paragraph{$\Delta$-layer configurations generalize across datasets, but affect overall accuracy.}
To study the sensitivity of accuracy to the choice of $\Delta$-layers, we evaluate DS-Qwen-1.5B and DS-Qwen-7B using 10 different $\Delta$-layer configurations on Mixed120. Figure~\ref{fig:ablation2} shows that overall accuracy can vary noticeably across configurations, indicating that selecting suitable $\Delta$-layers is important for each model. This effect is more visible for the larger model, where accuracy is higher and more sensitive to the underlying configuration. At the same time, the relative trends across datasets remain largely consistent for a given model as the $\Delta$-layer configuration changes. This suggests that, although the absolute accuracy depends on the selected $\Delta$-layers, a good configuration generalizes across datasets and does not exhibit dataset-specific behavior.

\subsection{Effect of Recency Window $L$}
\begin{figure}[htbp]
    \centering
    \includegraphics[width=\columnwidth]{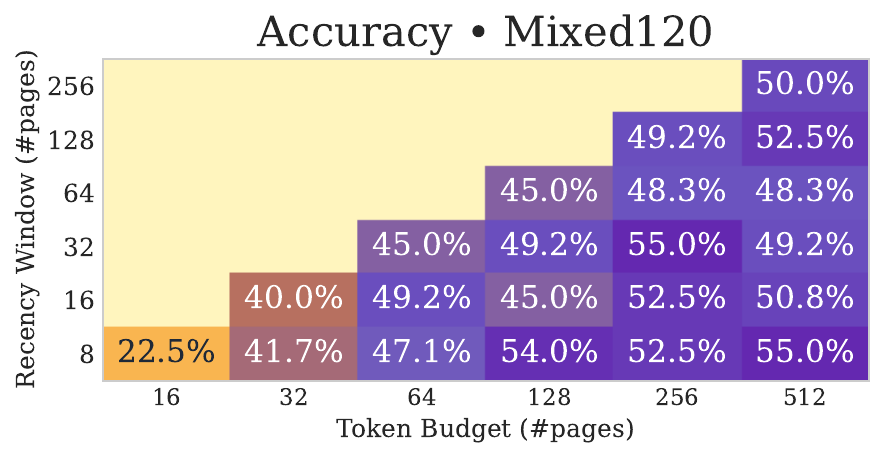}
    \caption{Accuracy under different page budgets $K$ and recency-window sizes $L$ for DS-Qwen-7B}
    \label{fig:ablation3}
\end{figure}

Figure~\ref{fig:ablation3} shows the effect of the recency-window size $L$ on accuracy under different page budgets $K$ for DS-Qwen-7B on Mixed120. We observe that accuracy is sensitive to the choice of $L$, with differences of up to 10\% across settings. Under larger page budgets, such as $K$$=$$256$ and $K$$=$$512$ pages (4k and 8k tokens), smaller recency windows tend to perform best, with $L$$=$$8$ giving the highest accuracy. In contrast, under the smallest budget, $K$$=$$64$ pages (1k tokens), a larger recency window performs better. This trend reflects a trade-off between preserving very recent context and allocating budget to a broader set of salient tokens: when the budget is large, the model benefits more from broader contextual coverage, whereas under tighter budgets, retaining recent context becomes more important.

\section{Related Work}
\paragraph{Efficient long-context inference.}
Long-context LLMs face quadratic compute and memory overhead from full self-attention, making inference increasingly dominated by KV-cache bandwidth. Even with optimized kernels such as FlashAttention \citep{dao2023flashattention} and paged caching \citep{kwon2023efficient}, decoding throughput scales poorly with sequence length. Modern architectures (e.g., Llama-3.1, GPT-4o, Claude 3.5 Sonnet) extend context to 128k–200k tokens through rotary positional encoding \citep{su2024roformer}, yet runtime remains bottlenecked by repeated KV reads rather than arithmetic compute, highlighting the need for structural sparsity that reduces redundant memory access.

\paragraph{Architectural and KV-compression methods.}
Architectural approaches such as Multi-Query and Grouped-Query Attention \citep{shazeer2019fast,ainslie2023gqa} reduce redundant KV heads, while recurrent alternatives like RWKV \citep{peng2023rwkv}, RetNet \citep{sun2023retentive}, and Mamba \citep{gu2023mamba} replace self-attention with stateful recurrence.
These designs improve efficiency but require model retraining and often underperform Transformers on complex reasoning tasks.
In contrast, KV-compression strategies optimize inference at runtime. Quantization \citep{xiao2023smoothquant,yao2022zeroquant,dettmers2022gpt3,liu2024kivi} lowers precision to save bandwidth, whereas pruning methods exploit sparsity to drop less important tokens. Eviction-based schemes such as H2O \citep{zhang2023h2o}, SnapKV \citep{li2024snapkv}, TOVA \citep{oren2024transformers}, ScissorHands \citep{liu2023scissorhands}, R-KV \citep{cai2025r}, and RaaS \citep{hu2025raas} bound memory by discarding low-score pages, but may lose tokens that later become critical for reasoning continuity.

\paragraph{Sparse and selection-based attention.}
Selection-based methods preserve the full KV cache but compute attention only over a subset of salient tokens. Early static patterns in Sparse Transformers, Longformer, and BigBird \citep{child2019generating,beltagy2020longformer,zaheer2020big} established fixed sparsity layouts, later refined into adaptive mechanisms guided by query-dependent importance.
Quest \citep{tang2024quest} scores KV pages against the current query to retrieve the most relevant subset, while TidalDecode \citep{yang2024tidaldecode} exploits the strong spatial coherence of attention across layers by performing full attention only in a few token-selection layers and reusing the selected tokens in intermediate sparse layers.
SeerAttention-R \citep{gao2025seerattention} employs a self-distilled gating module to learn block-sparse attention, achieving near-lossless decoding but requiring additional training.
However, existing sparse attention methods either incur notable accuracy degradation at low retention ratios or depend on costly post-training procedures to recover performance, both of which substantially increase decoding length and computational overhead for reasoning tasks.
In contrast, \textbf{DELTA} is proposed as a selection-based, training-free approach that leverages inter-layer attention correlation during reasoning to maintain high accuracy under reduced token budgets, without extending the overall generation length.

\section{Conclusion}
We introduced DELTA, a training-free, layer-aware sparse attention mechanism that improves the efficiency of long-context reasoning in large language models. By leveraging cross-layer correlation and gradual evolution of token importance, DELTA computes full attention only in a few key $\Delta$-layers and reuses their selected high-recall subsets across subsequent sparse attention layers. This design substantially reduces decoding-time bandwidth and latency while maintaining accuracy comparable to full attention. Experiments on multiple reasoning benchmarks confirm that DELTA achieves consistent speedups over state-of-the-art sparse and eviction-based methods without retraining, highlighting layer-aware reuse as a promising direction for efficient reasoning-time inference. 

\section*{Limitations}
While DELTA enables efficient long-context reasoning, it has the following limitations.
\paragraph{KV-memory footprint.}
DELTA preserves the full KV cache in HBM and reduces compute rather than peak memory. As a result, it does not directly address out-of-memory failures at extreme context lengths or on smaller GPUs. Future work includes integrating DELTA with complementary memory-saving techniques (e.g., quantization, eviction under guarantees, or offloading) while maintaining high selection recall.

\paragraph{Generality.}
Our results are limited to distilled DeepSeek-R1 models evaluated on reasoning benchmarks, mainly math and science QA. Generalization to other architectures, modalities, or workloads such as open-ended conversation and code generation remains unverified and may require re-tuning of the $\Delta$-layer schedule and context budgets.

\paragraph{Sensitivity and overhead.}
Performance depends on $\Delta$-layer placement and $(K,L)$; the max-attention scoring adds small overhead and can lag under fast attention drift. Adaptive per-sample scheduling or lightweight learned selectors are promising fixes.

\section*{Acknowledgments}
We sincerely thank all the reviewers for their time and constructive comments. This material is based upon work supported by  NSF award number  2224319, REAL@USC-Meta center, and VMware gift. 

% \section{Preamble}
% \subsection{Appendices}
% \section{Bib\TeX{} Files}

\bibliography{custom}

\clearpage
\appendix

\section{Algorithm}
\label{app:algorithm}

At each refresh layer, DELTA converts attention weights into token importance scores by taking the maximum attention over heads for each token. It then aggregates token scores within each page, always preserves the most recent $\ell=\lceil L/P\rceil$ pages, and fills the remaining budget with the highest-scoring older pages. The resulting page set is reused by subsequent sparse layers until the next refresh layer.

\begin{algorithm}
\small
\caption{DELTA decoding for one generation step}
\label{alg:delta}
\begin{algorithmic}[1]
\Require current hidden state $h_t^0$, full KV cache $\mathcal{C}_t$, warm-up depth $r$, refresh-layer set $\mathcal{D}$, page budget $k$, recency window $L$, page size $P$
\State $\rho \gets$ all pages in $\mathcal{C}_t$
\State $\ell \gets \lceil L / P \rceil$ \Comment{number of recency pages}
\For{$i = 1$ to $N$}
    \If{$i \le r$}
        \State Run layer $i$ with full attention over $\mathcal{C}_t$
        \State Obtain updated hidden state $h_t^i$
    \ElsIf{$i \in \mathcal{D}$}
        \State Run layer $i$ with full attention over $\mathcal{C}_t$
        \State Obtain updated hidden state $h_t^i$ and attention weights $A^i$
        \State $\rho \gets \textsc{RefreshPages}(A^i, k, \ell, P)$
    \Else
        \State Run layer $i$ with sparse attention over the selected pages $\rho$
        \State Obtain updated hidden state $h_t^i$
    \EndIf
\EndFor
\State \Return next-token prediction from $h_t^N$
\end{algorithmic}
\end{algorithm}

\begin{algorithm}
\small
\caption{Refreshing the selected pages in DELTA}
\label{alg:delta_refresh}
\begin{algorithmic}[1]
\Procedure{RefreshPages}{$A^i, k, \ell, P$}
    \State Convert the attention weights at layer $i$ into a token importance score
    \Statex \hspace{\algorithmicindent} by taking, for each token, its largest attention weight across heads
    \State Group tokens into pages of size $P$
    \State Compute a page score by summing the importance scores of tokens in the same page
    \State Keep the last $\ell$ pages to preserve recency
    \State From the remaining older pages, select the top-$(k-\ell)$ pages by page score
    \State Return the union of the recent pages and the selected high-score pages
\EndProcedure
\end{algorithmic}
\end{algorithm}

\section{Extended Experimental Results}
\subsection{Speedup Results}
\begin{figure}[htbp]
    \centering
    \includegraphics[width=\columnwidth]{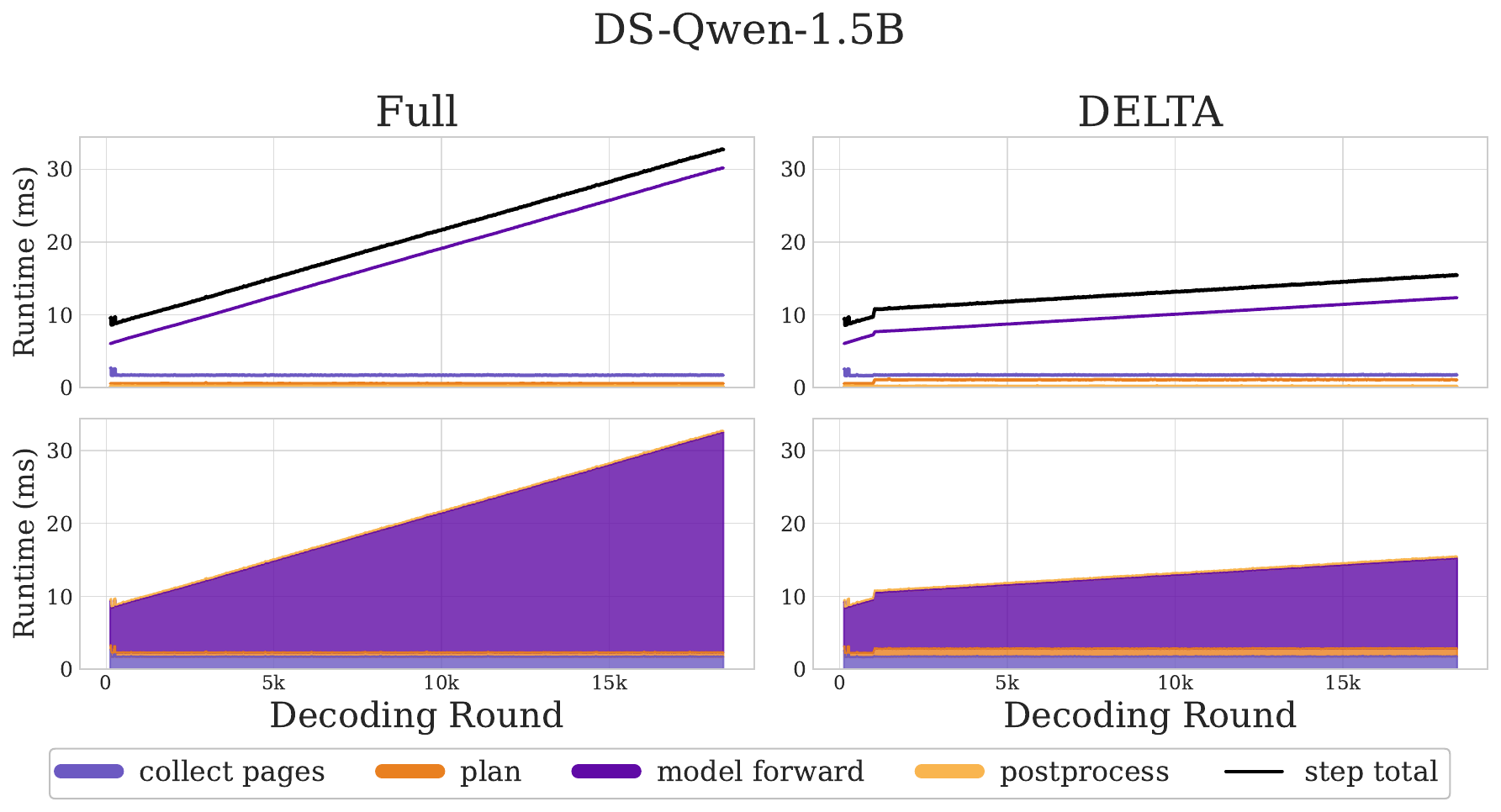}
    \caption{Runtime breakdown of DELTA and full attention on DS-Qwen-1.5B, corresponding to the speedup setting in Figure~\ref{fig:result2} (right). We report the latency of collect-pages, planning, model forward, post-processing, and total step time across decoding steps. DELTA introduces a small overhead for page collection and planning, but substantially reduces model-forward time, which dominates the overall runtime.}
    \label{fig:app:speedup1}
\end{figure}

\begin{figure}[htbp]
    \centering
    \includegraphics[width=\columnwidth]{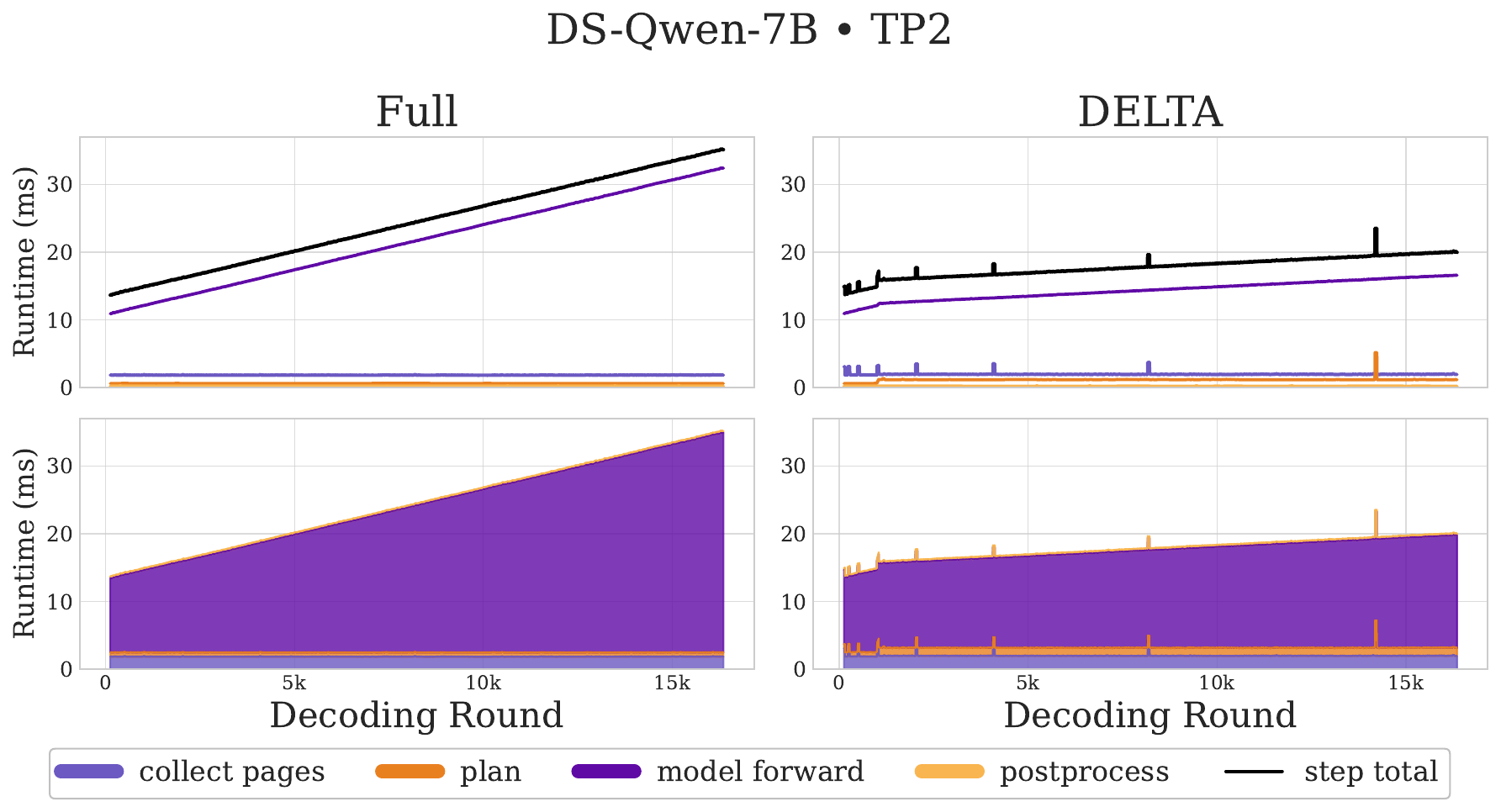}
    \caption{Runtime breakdown of DELTA and full attention on DS-Qwen-7B. As in DS-Qwen-1.5B, the additional overhead from collect-pages and planning remains small compared to the savings in model-forward time, leading to a clear reduction in total step latency.}
    \label{fig:app:speedup2}
\end{figure}

\begin{figure}[htbp]
    \centering
    \includegraphics[width=\columnwidth]{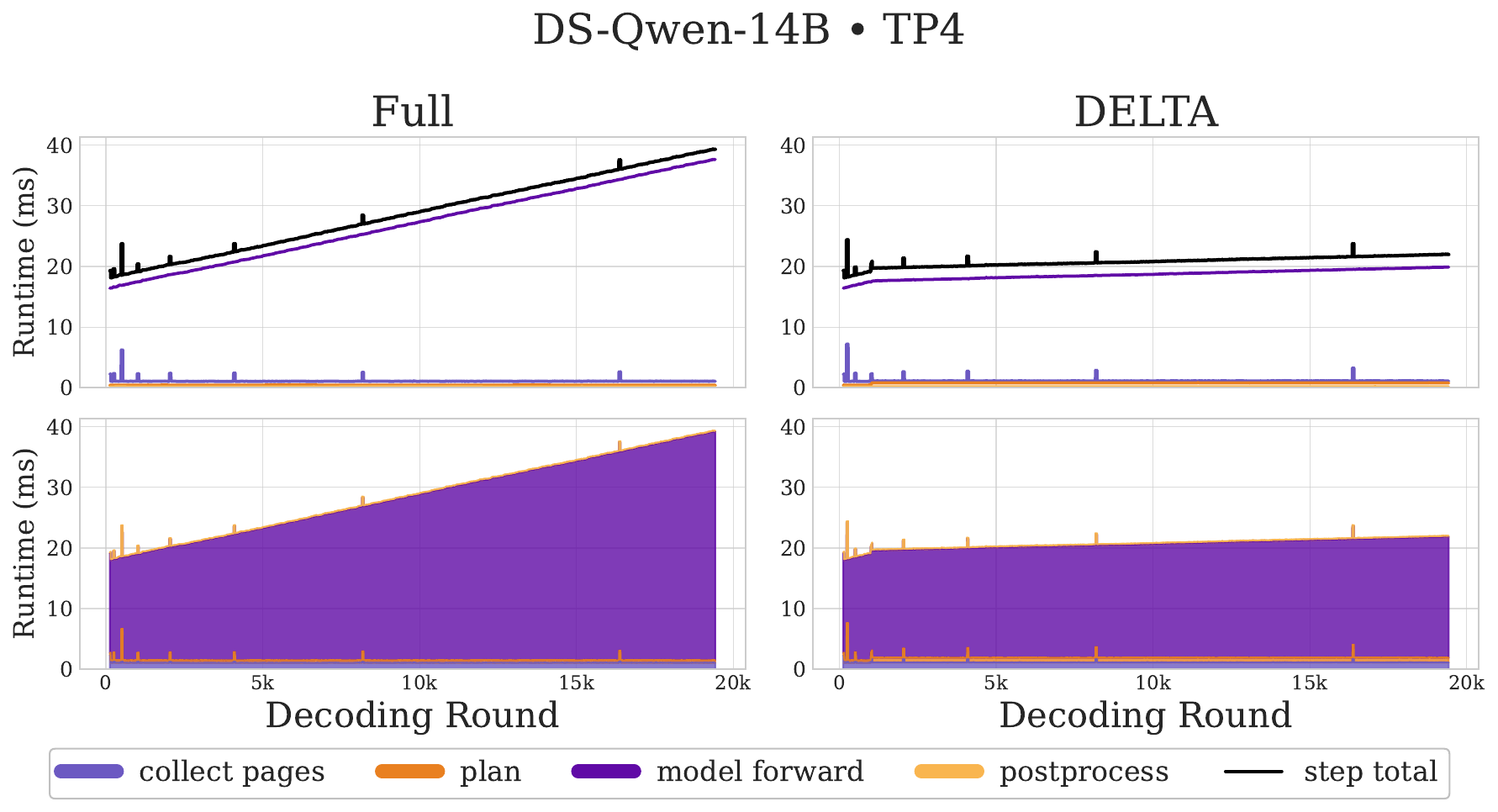}
    \caption{Runtime breakdown of DELTA and full attention on DS-Qwen-14B. The same trend holds at larger model scale: DELTA incurs modest non-forward overhead while consistently lowering forward latency, resulting in lower end-to-end decoding time per step.}
    \label{fig:app:speedup3}
\end{figure}

\paragraph{Experimental Setup.}
Figures~\ref{fig:app:speedup1}--\ref{fig:app:speedup3} present the per-decoding-round runtime breakdown for the speedup experiments in Figure~\ref{fig:result2} (right) for three model settings: DS-Qwen-1.5B on a single GPU, DS-Qwen-7B with tensor parallelism degree 2 (TP2, using two GPUs), and DS-Qwen-14B with tensor parallelism degree 4 (TP4, using four GPUs). In each case, the figure compares Full and DELTA with page budget $K$$=$$64$ side by side across decoding rounds, so the plots show not only the speed difference but also how the runtime composition evolves as generation proceeds and the effective context grows.

\paragraph{Breakdown Components.}
Each decoding step is decomposed into four measured components: \texttt{collect-pages}, \texttt{planning}, \texttt{model-forward}, and \texttt{post-processing}. The \texttt{collect-pages} term covers gathering the KV-page metadata and page indices needed for the current step. The \texttt{planning} term measures the preparation of the decode wrappers and related execution metadata. The \texttt{model-forward} term measures the actual decode computation, and for DELTA this also includes the planner-side attention dump, page-score computation, and top-$k$ page selection logic that are executed as part of the forward path. Finally, \texttt{post-processing} covers the remaining bookkeeping after the forward pass, such as extracting the next token and updating request state. The black \texttt{step total} curve is the sum of these components, while the stacked plots visualize how each component contributes to the total latency at each decoding round.

\paragraph{Interpretation.}
Compared with Full, DELTA introduces higher overhead in \texttt{collect-pages} and \texttt{planning} because it must support two attention paths rather than one: a full-context planner/dump attention path used to obtain attention statistics for page selection, and a subset-attention path used after the selected pages have been determined. Consequently, DELTA must gather page information and prepare execution state for both the full and subset paths, which increases these overhead terms. Nevertheless, the dominant effect in all three settings is the reduction in \texttt{model-forward} time. Although the DELTA \texttt{model-forward} component still includes the page-scoring and top-$k$ selection logic, it substantially reduces the later-layer attention workload by running those layers on a selected subset of KV pages instead of the full context. This difference becomes increasingly visible at later decoding rounds, where the cost of full-context attention continues to rise with sequence length, while DELTA keeps the later-layer attention cost much smaller, yielding consistently lower total decoding latency for DS-Qwen-1.5B, DS-Qwen-7B TP2, and DS-Qwen-14B TP4.

\section{Effect of Recency Window $L$, Per-dataset Results}
\begin{figure}[htbp]
    \centering
    \includegraphics[width=\columnwidth]{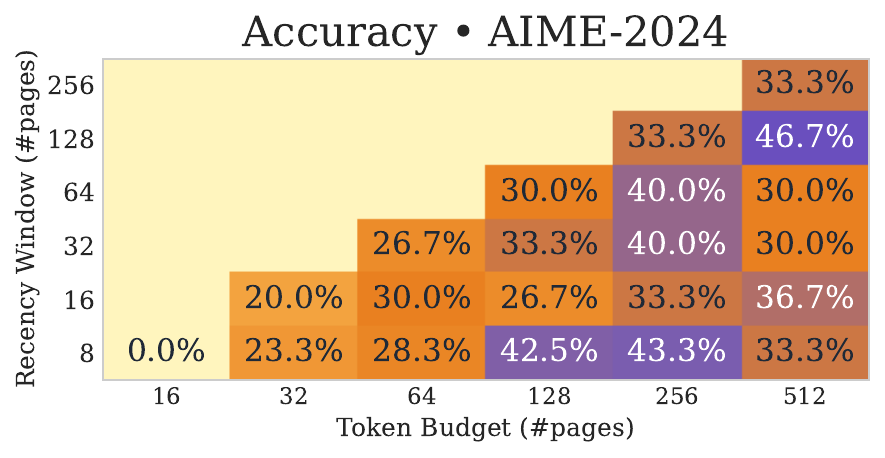}
    \caption{Accuracy on \textsc{AIME-2024} under different page budgets $K$ and recency-window sizes $L$ for DS-Qwen-7B. The best setting varies with budget: $L$$=$$8$ is best at $K$$=$$32,128,256$, $L$$=$$16$ is best at $K$$=$$64$, and the largest budget $K=512$ peaks at $L=128$, indicating a non-monotonic dependence on $L$.}
    \label{fig:app:heatmap1}
\end{figure}

\begin{figure}[htbp]
    \centering
    \includegraphics[width=\columnwidth]{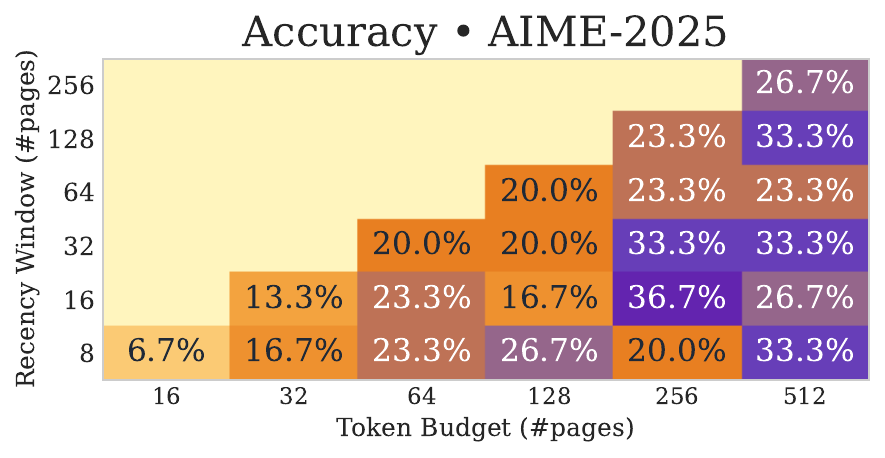}
    \caption{Accuracy on \textsc{AIME-2025} under different page budgets $K$ and recency-window sizes $L$ for DS-Qwen-7B. Smaller recency windows are strongest at budgets $K$$=$$32$--$128$, $L$$=$$16$ gives the best result at $K$$=$$256$, and several $L$ values tie at $K$$=$$512$, showing a mixed dependence on the recency window.}
    \label{fig:app:heatmap2}
\end{figure}

\begin{figure}[htbp]
    \centering
    \includegraphics[width=\columnwidth]{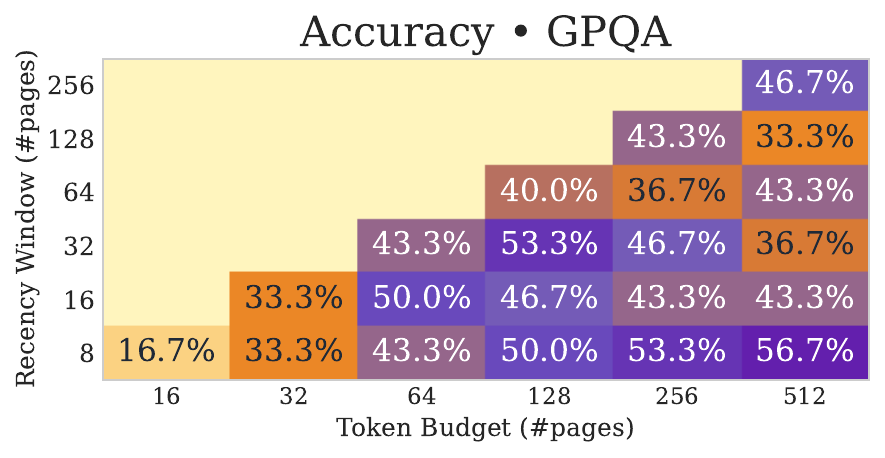}
    \caption{Accuracy on \textsc{GPQA} under different page budgets $K$ and recency-window sizes $L$ for DS-Qwen-7B. The preferred recency window shifts with budget: the best accuracy occurs at $L$$=$$16$ for $K$$=$$64$, at $L$$=$$32$ for $K$$=$$128$, and at $L$$=$$8$ for $K$$=$$256$ and $K$$=$$512$, indicating that the balance between recent-context preservation and broader page coverage matters for scientific reasoning.}
    \label{fig:app:heatmap3}
\end{figure}

\begin{figure}[htbp]
    \centering
    \includegraphics[width=\columnwidth]{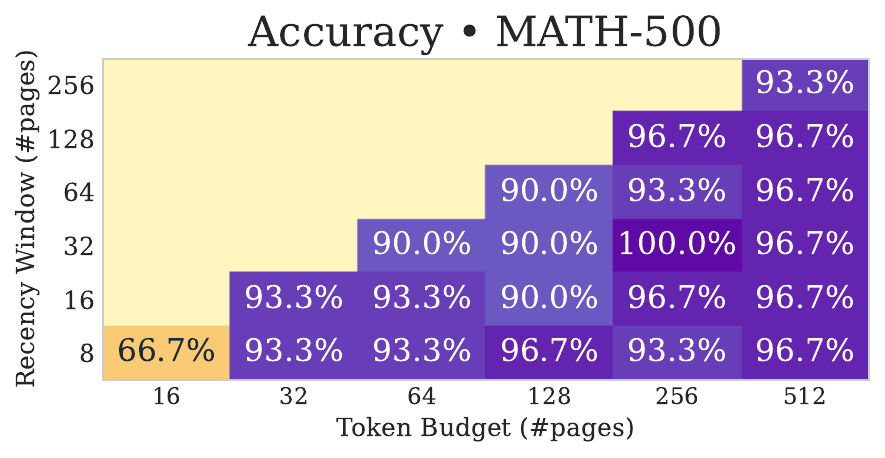}
    \caption{Accuracy on \textsc{MATH-500} under different page budgets $K$ and recency-window sizes $L$ for DS-Qwen-7B. Compared with the other datasets, performance is relatively insensitive to $L$ across most budgets, although $K$$=$$256$ reaches its peak at $L$$=$$32$.}
    \label{fig:app:heatmap4}
\end{figure}

Figures~\ref{fig:app:heatmap1}--\ref{fig:app:heatmap4} show the per-dataset version of Figure~\ref{fig:ablation3}, breaking down the effect of recency-window size $L$ across \textsc{AIME-2024}, \textsc{AIME-2025}, \textsc{GPQA}, and \textsc{MATH-500}. Only valid configurations with $L < K$ are shown. The dependence on $L$ is visible in all four datasets, but the pattern is not uniform. On \textsc{AIME-2024}, small or moderate recency windows are usually best at intermediate budgets, while the best setting shifts to a larger $L$ at the largest budget. On \textsc{AIME-2025}, smaller $L$ values perform best at budgets $K=32$--$128$, while a moderate recency window performs best at $K=256$. On \textsc{GPQA}, the preferred $L$ changes with budget, indicating a clearer trade-off between preserving recent context and retaining broader coverage. In contrast, \textsc{MATH-500} is comparatively insensitive to $L$ across most budgets, with only localized gains from particular settings. Overall, these per-dataset results show that the effect of $L$ is dataset- and budget-dependent rather than following a single uniform trend.

\section{Consecutive-Layer Attention Shift}
\label{sec:app:attention-similarity}

\begin{figure*}[t!]
    \centering
    \includegraphics[width=1\linewidth]{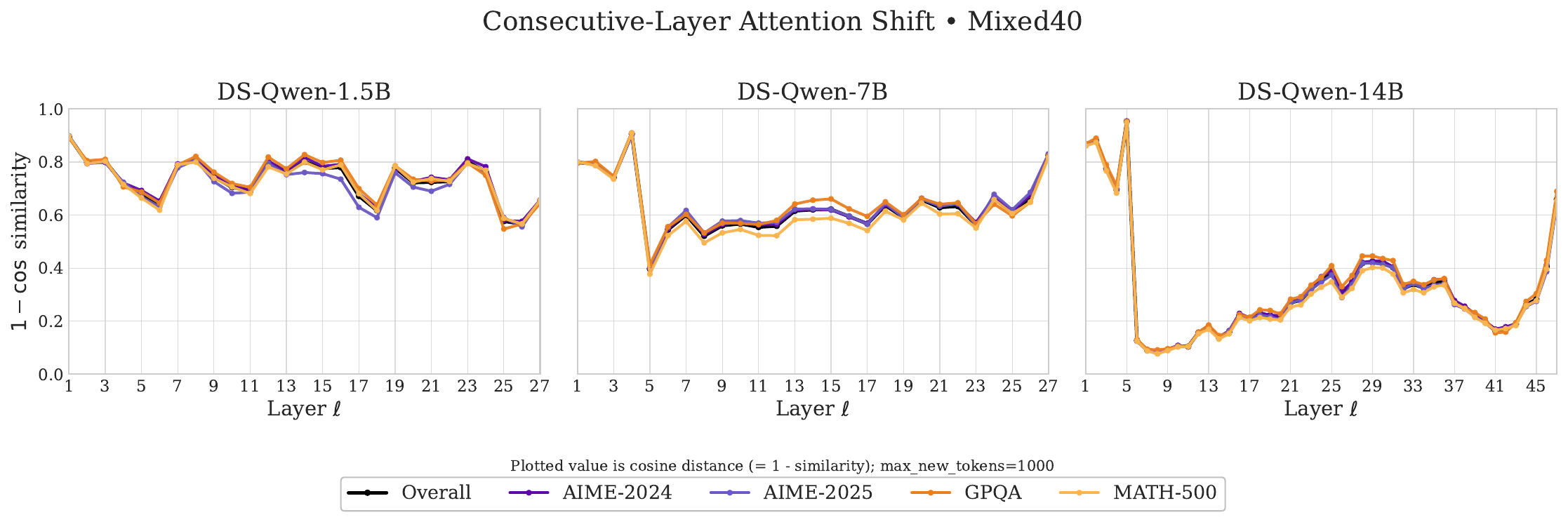}
    \caption{Attention shift between consecutive layers, measured as $1-\text{cosine similarity}$ between their attention distributions during autoregressive decoding. Each panel shows a different model, with the black curve denoting the overall mean on \texttt{Mixed40} and the colored curves showing per-dataset means. Peaks indicate layers where attention behavior changes sharply relative to the previous layer, while low-shift regions indicate smoother transitions. The major peaks are largely consistent across datasets within each model, suggesting that consecutive-layer attention shift is primarily driven by model architecture and provides a useful signal for selecting $\Delta$-layers.}
    \label{fig:app:attention_similarity}
\end{figure*}

\paragraph{Setup.}
Figure~\ref{fig:app:attention_similarity} plots the attention shift between consecutive layers as a function of layer index for three models: DS-Qwen-1.5B, DS-Qwen-7B, and DS-Qwen-14B, without tensor parallelism. The probe dataset is a fixed \texttt{Mixed40} artifact constructed from the \texttt{Mixed120} dataset by selecting exactly $10$ prompts from each of \textsc{AIME-2024}, \textsc{AIME-2025}, \textsc{GPQA}, and \textsc{MATH-500}, for a total of $40$ prompts. Each prompt is decoded for up to $1000$ generated tokens.

\paragraph{Measurement Procedure.}
At each one-token autoregressive decode step, and for each transformer layer $\ell$, we capture the attention distribution of the last query token over the valid context positions. Concretely, we collect the attention weights for all heads, apply a softmax over the context dimension, and flatten the resulting per-head maps into a single vector $a_{\ell}^{(t)}$ for decode step $t$. We first compute cosine similarity between consecutive layers,
\[
s_{\ell-1,\ell}^{(t)}
=
\frac{a_{\ell-1}^{(t)} \cdot a_{\ell}^{(t)}}
{\|a_{\ell-1}^{(t)}\| \, \|a_{\ell}^{(t)}\|},
\]
and then define the plotted quantity as the corresponding attention shift,
\[
d_{\ell-1,\ell}^{(t)} = 1 - s_{\ell-1,\ell}^{(t)}.
\]
For each layer pair, we report the mean shift over all valid one-token decode steps and prompts. Thus, larger values correspond to larger changes in attention behavior from the previous layer.

\paragraph{Figure Structure.}
In each panel, the black curve shows the overall mean attention shift aggregated over all $40$ prompts in \texttt{Mixed40}, while the colored curves show the per-dataset means for \textsc{AIME-2024}, \textsc{AIME-2025}, \textsc{GPQA}, and \textsc{MATH-500}. The horizontal axis is the layer index $\ell$, corresponding to the right member of the consecutive pair $(\ell-1,\ell)$, and the vertical axis is the mean value of $1-\text{cosine similarity}$ between the attention distributions of those two neighboring layers. High values therefore mark layers whose attention pattern changes sharply relative to the previous layer, whereas low values indicate smoother transitions and more similar consecutive-layer behavior.

\paragraph{Overall Trends Across Models.}
The three models exhibit substantially different shift scales. Averaged over all consecutive layer pairs, the overall shift is approximately $0.733$ for DS-Qwen-1.5B, $0.631$ for DS-Qwen-7B, and $0.323$ for DS-Qwen-14B. Thus, the larger model shows much smaller consecutive-layer shifts on average, indicating more aligned attention behavior across neighboring layers. At the same time, the shift is not monotonic with depth in any model. Instead, all three curves contain localized peaks that mark abrupt transitions in attention behavior. DS-Qwen-7B shows a particularly strong peak between layers $(3,4)$ with shift $\approx 0.907$, followed immediately by a much smaller shift between $(4,5)$ of about $0.397$. DS-Qwen-14B shows its strongest peak between $(4,5)$ with shift $\approx 0.953$, after which a long middle block has very small shifts; for example, pairs such as $(6,7)$, $(7,8)$, and $(8,9)$ are all near $0.10$ or below. DS-Qwen-1.5B maintains comparatively high shift throughout much of the stack, with its largest shift appearing at the earliest pair $(0,1)$ at about $0.896$.

\paragraph{Dataset-wise Behavior.}
The per-dataset curves generally track the black overall curve closely within each model. This indicates that the dominant layerwise shift structure is driven more by model architecture than by the particular reasoning dataset. Quantitatively, the deviation from the overall curve is modest: the mean absolute difference between a dataset-specific curve and the overall curve is typically on the order of $0.004$--$0.019$, depending on the model and dataset. \textsc{GPQA} and \textsc{MATH-500} show somewhat larger deviations than the AIME subsets in some settings, but the major transition layers remain in essentially the same locations. In other words, the datasets affect the magnitude of the shift curve more than its global structure.

\paragraph{Interpretation.}
These results suggest that consecutive-layer attention behavior is organized into stages rather than changing smoothly and uniformly across depth. Peaks in the shift curve mark layers where the model substantially changes how it distributes attention over the context, while low-shift regions indicate stretches of neighboring layers with more similar attention behavior. This interpretation is especially relevant for DELTA: layers with large shift are natural candidates for $\Delta$-layers, since they mark points where the model’s attention behavior changes most sharply relative to the previous layer. Conversely, regions with consistently low shift suggest groups of layers with more redundant attention behavior, which are more natural candidates for stronger compression or subset-based treatment. Under this view, the probe provides an architectural signal for identifying promising DELTA layer placements by highlighting where the model undergoes its strongest layer-to-layer attention shift.

\begin{figure*}[t!]
    \centering
    \includegraphics[width=1\linewidth]{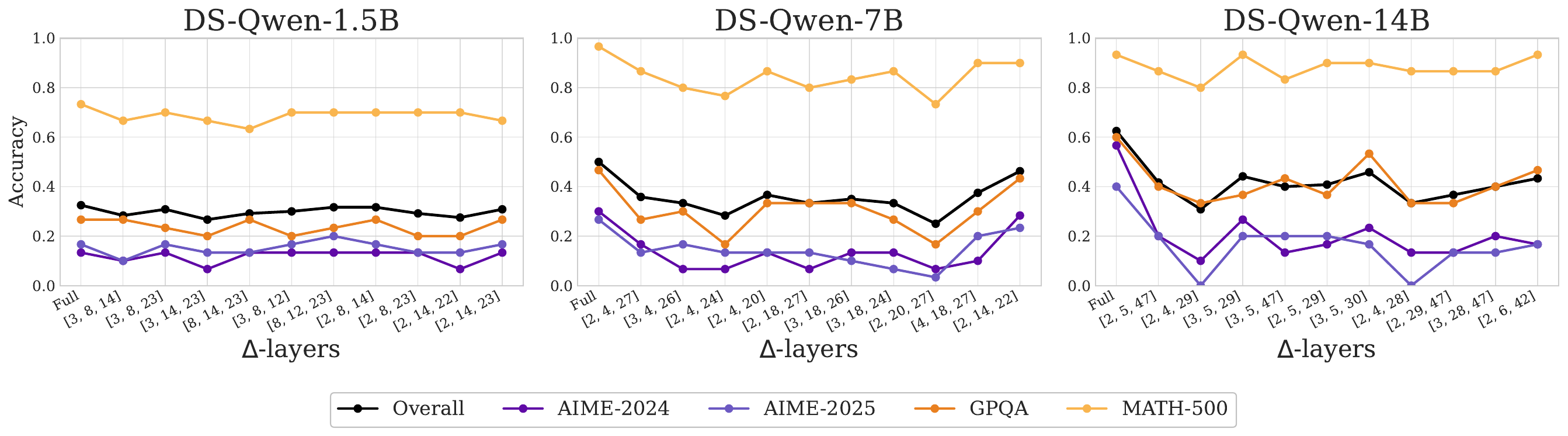}
    \caption{Accuracy across different $\Delta$-layer configurations on Mixed120 for DS-Qwen-1.5B, DS-Qwen-7B, and DS-Qwen-14B. Each panel reports overall accuracy and the per-dataset breakdown on \textsc{AIME-2024}, \textsc{AIME-2025}, \textsc{GPQA}, and \textsc{MATH-500}.}
    \label{fig:app:fig6_accuracy_three_models}
\end{figure*}

\subsection{Per-model Accuracy Under Different $\Delta$-layer Configurations}
\paragraph{Extended results across three model sizes.}
Figure~\ref{fig:app:fig6_accuracy_three_models} extends Figure~\ref{fig:ablation2} by including DS-Qwen-14B. To construct the $\Delta$-layer configurations evaluated here, we use the prominent shift points identified in the consecutive-layer attention-shift plots from Section~\ref{sec:app:attention-similarity}, selecting candidate layers around these peaks while ensuring coverage across network depth. Across all models, overall accuracy varies across $\Delta$-layer configurations, showing that $\Delta$-layer placement is important. This effect is stronger for larger models, where accuracy is more sensitive to the chosen configuration. However, within each model, the relative trends across datasets remain broadly consistent. This suggests that although the absolute accuracy depends on the selected $\Delta$-layers, good configurations generalize across datasets rather than being dataset-specific.

\section{Breakdown of DELTA Runtime Overhead}
To better understand the runtime overhead introduced by DELTA, we conduct a component-wise microbenchmark that isolates the decode-time attention kernel and the two downstream primitives used for page-aware selection. Specifically, on a single GPU without tensor parallelism, we measure four components: (1) the baseline FlashInfer decode attention kernel, (2) a JIT-instrumented FlashInfer variant that additionally dumps attention logits, (3) DELTA’s fused page-scoring kernel, which aggregates token-level attention into page-level importance scores, and (4) DELTA’s page-selection step, which selects the pages retained for sparse attention.

\begin{figure}
    \centering
    \includegraphics[width=1\linewidth]{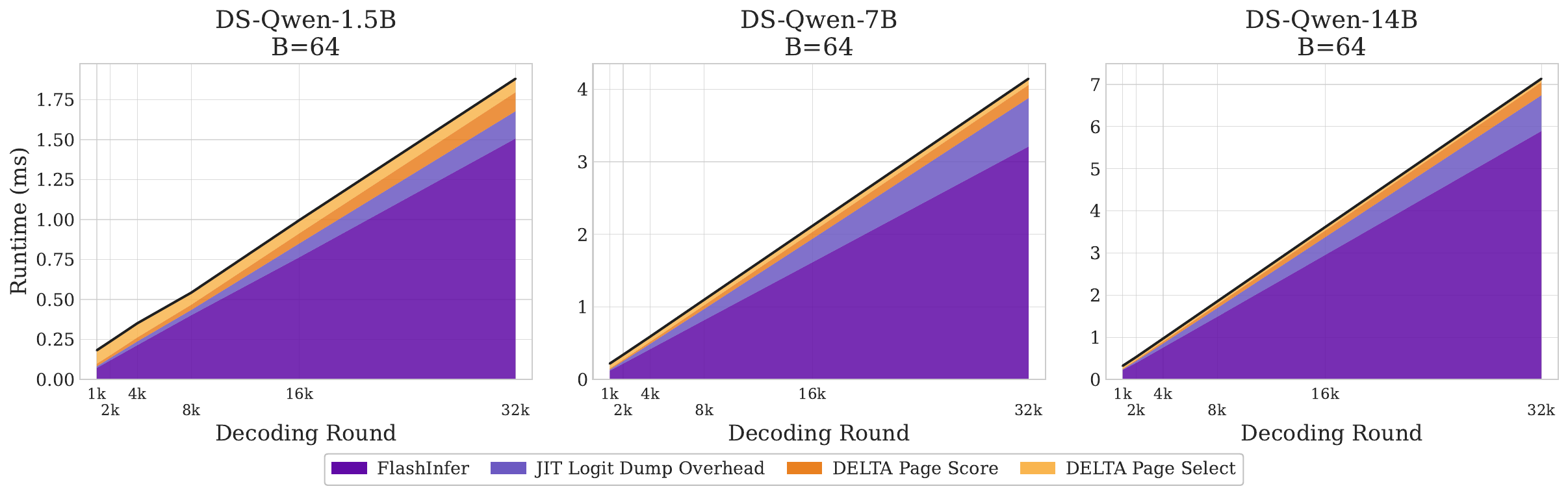}
    \caption{Stacked runtime breakdown of DELTA’s auxiliary overhead across decoding rounds on a single GPU ($B=64$). The areas show baseline FlashInfer attention, incremental JIT logit-dump overhead, DELTA page scoring, and DELTA page selection. The relative overhead is highest at short contexts due to the nearly fixed cost of page selection, but decreases as context length grows and the baseline attention kernel becomes dominant.}
    \label{fig:app:fig8_kernel_runtime_comparison}
\end{figure}

Figure~\ref{fig:app:fig8_kernel_runtime_comparison} shows the resulting runtime breakdown. The bottom area corresponds to the baseline FlashInfer kernel, the second area to the incremental overhead of JIT logit dumping, and the top two areas to DELTA’s page scoring and page selection. The results show that DELTA’s relative overhead is highest at short contexts, where page selection contributes an almost fixed per-step cost while the baseline attention kernel is still small. At 1k context, the total overhead is about 154\%, 81\%, and 42\% of the baseline FlashInfer runtime for the 1.5B, 7B, and 14B models, respectively. As context grows, the baseline attention cost increases much faster than the fixed portion of DELTA’s control path, so the relative overhead decreases. By 32k context, the total overhead drops to about 25\%, 29\%, and 21\% of the baseline runtime for the 1.5B, 7B, and 14B models, respectively.

The breakdown also reveals a shift in which component dominates. At 1k context, page selection is the largest source of overhead, accounting for roughly 68--76\% of the total extra cost across models. At longer contexts, the JIT logit-dump path becomes dominant, contributing roughly 46--72\% of the overhead at 32k, while fused page scoring remains smaller and page selection stays nearly flat at around 0.08--0.09\,ms. Overall, these results show that DELTA’s overhead is most pronounced at short decoding lengths, where fixed control costs are less amortized, and becomes progressively less significant as context grows. This trend is favorable for DELTA’s target setting, since long-context decoding is precisely where reducing attention cost matters most.

\end{document}